\documentclass{article}

\usepackage[preprint, nonatbib]{neurips_2021}

\usepackage[utf8]{inputenc} 
\usepackage[T1]{fontenc}    
\usepackage{hyperref}       
\usepackage{url}            
\usepackage{booktabs}       
\usepackage{amsfonts}       
\usepackage{nicefrac}       
\usepackage{microtype}      
\usepackage{xcolor}         
\usepackage{amsmath}
\usepackage{amssymb}
\usepackage{amsthm}
\usepackage[caption=false]{subfig}
\usepackage{tikz, pgfplots}
\pgfplotsset{compat=1.17}
\newtheorem{definition}{Definition}
\newtheorem{theorem}{Theorem}
\newtheorem{lemma}{Lemma}
\title{Privacy-Preserving Kickstarting Deep Reinforcement Learning with Privacy-Aware Learners}

\author{%
  Parham Gohari\\
  University of Texas at Austin\\
  Austin, TX 78712 \\
  \texttt{pgohari@utexas.edu} \\
  \And
   Bo Chen\\
   University of Florida \\
   Gainesville, Florida 32611 \\
   \texttt{bo.chen@ufl.edu} \\
   \AND
   Bo Wu \\
   University of Texas at Austin\\
  Austin, TX 78712 \\
   \texttt{bowu86@gmail.com} \\
   \And
   Matthew Hale \\
   University of Florida \\
   Gainesville, Florida 32611 \\
   \texttt{matthewhale@ufl.edu} \\
   \And
   Ufuk Topcu \\
   University of Texas at Austin\\
  Austin, TX 78712 \\
   \texttt{utopcu@utexas.edu} \\
}

\begin{document}

\maketitle

\begin{abstract}
Kickstarting deep reinforcement learning algorithms facilitate a teacher-student relationship among the agents and allow for a well-performing teacher to share demonstrations with a student to expedite the student's training.
However, despite the known benefits, the demonstrations may contain sensitive information about the teacher's training data and existing kickstarting methods do not take any measures to protect it.
Therefore, we use the framework of differential privacy to develop a mechanism that securely shares the teacher's demonstrations with the student.
The mechanism allows for the teacher to decide upon the accuracy of its demonstrations with respect to the privacy budget that it consumes, thereby granting the teacher full control over its data privacy.
We then develop a kickstarted deep reinforcement learning algorithm for the student that is privacy-aware because we calibrate its objective with the parameters of the teacher's privacy mechanism.
The privacy-aware design of the algorithm makes it possible to kickstart the student's learning despite the perturbations induced by the privacy mechanism.
From numerical experiments, we highlight three empirical results: (i) the algorithm succeeds in expediting the student's learning, (ii) the student converges to a performance level that was not possible without the demonstrations, and (iii) the student maintains its enhanced performance even after the teacher stops sharing useful demonstrations due to its privacy budget constraints. 
\end{abstract}

\section{Introduction} \label{sec:intro}
Neural networks, widely used in deep reinforcement learning, are known to be vulnerable to privacy attacks that target their training data.
For example, membership attacks can determine whether a specific data point has been used to train the neural network under attack or not \cite{shokri2017membership,truex2019demystifying,gomrokchi2019privattack,pan2019you}.
Such vulnerabilities risk the privacy of users who provide personal information such as medical data or location data to train deep reinforcement learning agents.
For example, in the emerging medical applications of deep reinforcement learning such as dynamic treatment recommendation \cite{wang2018supervised}, lack of privacy measures could breach the privacy of the involved patients.
Therefore, rigorous privacy guarantees are needed when training deep reinforcement learning agents with personal and sensitive user information.

We consider the problem of \textit{kickstarting} deep reinforcement learning through a privacy lens.
Kickstarting methods employ already trained agents as teachers to share demonstrations with a student who has to learn the same task from scratch.
Existing methods in \cite{parisotto2015actor, rusu2015policy, schmitt2018kickstarting} can significantly expedite the training of the students; however, they do not take any privacy measures in sharing the teacher demonstrations.
Due to the privacy vulnerabilities of neural networks, there is a need to retain the benefits of kickstarting while ensuring that the teacher's sensitive information is kept private.
To this end, we make two contributions to redesign the students and the teachers, and develop a novel privacy-preserving kickstarting algorithm.

In the first contribution, we use the framework of differential privacy \cite{dwork2014algorithmic} to protect the teacher's privacy while sharing its policy with the student.
Through differential privacy's quantitative formulation, it is possible to provide rigorous privacy guarantees, which is one of the main reasons that it is widely used to protect the privacy of the users of machine learning algorithms \cite{10.1145/2976749.2978318,papernot2016semi}.
From a differential privacy perspective, the policy that the teachers share with the students is a query of their sensitive training data, and we will randomize policies to protect them.

We use the Dirichlet mechanism, which was shown to obfuscate small differences between two entries of a finite-support probability vector \cite{9328150} to enforce differential privacy.
Our main contribution is to provide new analyses that generalize the Dirichlet mechanism's privacy protections to all entries of a probability vector.
A policy is a probability distribution over the action space of the environment.
Assuming that the environment's action space is finite, we deploy the Dirichlet mechanism as the privacy mechanism of the teachers, and our new analysis shows that the entire policy is protected.

We derive analytical expressions for the level of the Dirichlet mechanism's differential privacy and allow the teachers to enforce their desired strength of privacy protections, thereby granting them full control over their privacy.
Furthermore, the teachers can manage their \textit{privacy budget} throughout the sequence of demonstrations that they share with the students by keeping track of their history of previous privacy levels,.
A privacy budget measures the privacy loss due to repeated queries on a fixed dataset and we follow the convention to permit queries up to a budget and then to disallow queries once the budget is expended \cite{abadi2016deep,dankar2013practicing}.

In the second contribution, we design the student's learning algorithm.
The Dirichlet mechanism, similar to conventional differential privacy mechanisms, enforces privacy by perturbing its output which can be harmful to students who do not handle these outputs properly.
We design the students such that their objectives incentivize them to approximate the policy that the privacy-preserving demonstrations provide; however, we make sure that the students do not attempt to approximate the teacher's perturbed policy beyond an accuracy threshold that we determine from a concentration bound on the output of the Dirichlet mechanism.
The resulting privacy-aware students avoid being misled by the perturbations by automatically balancing the amount of information that they use in their learning from the demonstrations and their own interactions with the environment.

In the experiments, we implement the proposed privacy-preserving kickstarting algorithm using teachers with varied privacy budgets.
Across all privacy budgets tested, we observed that privacy-aware students converge roughly 4-times faster than their teachers and often outperform them even though their networks have $50\%$ smaller hidden layers.
However, teachers with higher privacy budgets train students with a superior performance level and robustness to environmental changes, which we measure with the run-to-run variance across different environment seeds.  
The performance level of the students who blindly follow the demonstrations are significantly worse than the privacy-aware students, indicating that the students' privacy-awareness is crucial to the utility of the privacy-preserving kickstarting algorithm.


\section{Preliminaries}\label{sec:prelim}
In this section, we will set the notation that we use throughout the the paper and formally define differential privacy. 
\subsection{Notation}
We denote by $\mathbb{N}$, $\mathbb{R}$, and $\mathbb{R}_+$ the set of positive integers, real numbers, and non-negative reals, respectively. For all $n\in\mathbb{N}$, let $[n] := \{1,\dots,n\}$. For a fixed $\eta\in\mathbb{R}_+$, we define the $\eta$-restricted simplex as
\begin{equation}\label{def:simplex}
    \Delta_{n,\eta}:=\left\{x\in\mathbb{R}^n \mid \sum\limits_{i\in[n]} x_i = 1, x_i\ge \eta, \forall{i}\in[n]\right\}.
\end{equation}
We use notation $\Delta_n$ to refer to the entire unit simplex, \textit{i.e.}, when $\eta=0$ in (\ref{def:simplex}).
Finally, we denote the $p$-norm of a vector $x\in\mathbb{R}^n$ by $\|x\|_p$.

\subsection{Problem Statement}
Deep reinforcement learning agents interact with an environment that is modeled as a Markov decision process comprising a state and action space, transition probabilities, and a reward function. 
We consider deep reinforcement learning agents who do not know the transition probabilities and must compute a policy that maximizes their cumulative rewards.
A policy is a probability distribution over the action space at every state of the environment.
We assume that the agents observe the environment's state space through $d$-dimensional feature vectors and we assume that the action space is finite.
As a result, the policy of the agents are members of the unit simplex.

\begin{figure}[t]
    \centering
            \tikzstyle{block} = [rectangle, draw, 
            text width=8em, text centered, rounded corners, minimum height=2.5em]
        \tikzstyle{line} = [draw, -latex]

    \begin{tikzpicture}[ node distance = 3.5em, auto, thick]
        \node [block](Teacher) {Teacher};
        \node [block, below of = Teacher](Student) {Student};            
        \node [block, below of= Student] (Environment) {Environment};        
        \path [line, text width = 3cm, align = center] (Student.0) --++ (5em,0em) |- node [near start]{\small{Student's Policy \\ $\pi_\text{student}(X_t)$}} (Environment.0);
        \path [line, text width = 4cm, align = center] (Environment.180) --++ (-5em,0em) |- node [near start]{\small{Student's Observation $X_t$,\\ Reward $r_t$}} (Student.180);
        \path [line] (Student.170) --++ (-3.5em,0em) |- node [near start]{\small{Student's Observation $X_t$}} (Teacher.180);
        \path [line, text width = 4cm, align = center] (Teacher.0) --++ (3.5em,0em) |- node [near start]{{\small{Teacher's Policy \\ with Privacy \\ Protections $\pi_{\text{teacher}}(X_t)$}}} (Student.10);
    \end{tikzpicture}
    \caption{Schematic of a privacy-preserving kickstarting algorithm. At every time step, the student interacts with the environment and has the option to ask for the teacher's policy. We use a privacy mechanism before sharing the teacher's policy to protect the privacy of the training dataset with which the teacher has been trained.}
    \label{fig:my_label}
\end{figure}
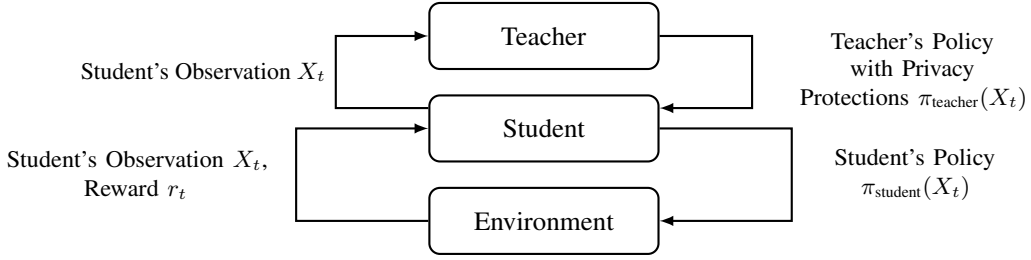

In kickstarting deep reinforcement learning, the agents have the option of asking for demonstrations from some already trained teacher agents as shown in Figure \ref{fig:my_label}. Precisely, at every state observation, the students can ask for the teachers' policy corresponding to the student's current environment observation. 
The teachers generate their policies using neural networks that take as input an environment observation and output a policy.
Due to the privacy vulnerabilities of neural networks, we pose the problem of privacy-preserving kickstarting, wherein we seek to find a privacy mechanism for the teachers to securely share their policies with the student and to find a deep reinforcement learning algorithm for the students to be able to benefit from such policies.

\subsection{Differential Privacy}

We use the notion of differential privacy \cite{dwork2014algorithmic} as the underlying definition of data privacy. Differential privacy is provided by a probabilistic mapping, or a mechanism, and requires the mechanism to generate approximately indistinguishable outputs for the inputs that are \textit{similar}.
We next define adjacency, which captures the notion of similarity.
\begin{definition}[\textsc{adjacency relationship}]\label{def:adjacency}
Let $X$ and $X'$ be two vectors in $\mathbb{R}^d$ and fix a constant $b\in\mathbb{R}_+$. Then $X$ and $X'$ are $b$-adjacent if and only if $\|X-X'\|_2 \le b$.
\end{definition}

The general definition of differential privacy allows for any adjacency relationship that the mechanism designer sees fit for its application. 
In kickstarting deep reinforcement learning, teachers share the policy that corresponds to the student's environment observation.
With environment observations as $d$-dimensional vectors, we define the adjacency relationship over the space of vectors in $\mathbb{R}^d$.
In Section \ref{sec:limitations}, we discuss the consequences of defining adjacency relationship according to Definition \ref{def:adjacency} and discuss how it can protect the privacy of the training datasets of the teachers.
Reflecting the definition of differential privacy in \cite{le2013differentially}, we now define $(\epsilon,\delta)$-differential privacy.

\begin{definition}[\textsc{$\mathbf{(\epsilon,\delta)}$-differential privacy}]\label{def:differential privacy}
Fix a probability space $(\Omega, \mathcal{F}, \mathbb{P})$, input space $D$, and a measurable output space $(R,\mathcal{M})$, where $\mathcal{M}$ is some $\sigma$-algebra. For constants $\epsilon\in\mathbb{R}_+$, $\delta\in[0,1)$, and $b\in\mathbb{R}_+$, a mechanism $M: \Omega \times D \mapsto R$ is $(\epsilon, \delta)$-differentially private if, for all measurable subsets $S\subseteq R$ and all $b$-adjacent $X$ and $X'$,
\begin{equation}
    \mathbb{P}\left(M(X)\in S\right) \le \exp(\epsilon) \cdot \mathbb{P}\left(M\left(X'\right)\in S\right) + \delta.
\end{equation}
\end{definition}

The differential privacy mechanism that we design for the teacher has its domain space $D$ set to $\mathbb{R}^d$, the space of environment observations.
The mechanism must share a policy with the student; therefore, the output space $R$ is the space of probability vectors over the environment's action space which we assume to be finite.
Fixing $m$ as the number of actions, the output space of the privacy mechanism is the $m$-fold simplex $\Delta_m$.
Later in the proofs we use the notion of \textit{probabilistic} differential privacy, which we define as follows:
\begin{definition}[\textsc{probabilistic differential privacy}]
Fix a probability space $(\Omega, \mathcal{F}, \mathbb{P})$, input space $D$, and a measurable output space $(R,\mathcal{M})$, where $\mathcal{M}$ is some $\sigma$-algebra. For constants $\epsilon\in\mathbb{R}_+$, $\delta\in[0,1)$, and $b\in\mathbb{R}_+$, a mechanism $M: \Omega \times D \mapsto R$ satisfies $(\epsilon, \delta)$-probabilistic differential privacy if we can partition $R$ into disjoint sets $R_1$ and $R_2$ such that, for all $X\in D$ and all measurable $S\subseteq R_1$,
\begin{equation}
    \mathbb{P}\left(M(X)\in S \right) \le \delta
\end{equation}
and, for all $X'\in D$ $b$-adjacent to $X$ and all measurable $S\subseteq R_2$,
\begin{equation} \label{eq: prob DP epsilon}
    \mathbb{P}\left(M(X)\in S\right) \le \exp(\epsilon) \cdot \mathbb{P}\left(M\left(X'\right)\in S\right).
\end{equation}
\end{definition}
We show that the privacy mechanism that we develop in the sequel is probabilistically differentially private.
We then use the fact that $(\epsilon,\delta)$-probabilistic differential privacy implies $(\epsilon,\delta)$-differential privacy \cite{machanavajjhala2008privacy} and show that the mechanism is differentially private according to Definition \ref{def:differential privacy}.

\section{Privacy-Preserving Teachers} \label{sec:teacher}


In this section, we develop the teacher's privacy mechanism to share its policy with the students.
In environments with finite action spaces, the policy is a vector within the unit simplex and privatizing the policy requires a privacy mechanism that does not destroy their special structure, namely the properties that all entries are non-negative and that their sum is 1.
Three common differential privacy mechanisms are the Gaussian, Laplacian, and exponential mechanism \cite{dwork2014algorithmic}.
The first two mechanisms inject infinite-support noise and break the structure of the policy, and the exponential mechanism has a finite output range whereas the teacher has infinitely many policies that belong to the unit simplex; therefore, they are not readily applicable to privatizing the teacher's policy in kickstarting.

The Dirichlet mechanism, first introduced in \cite{9328150}, can mask small differences in two arbitrary entries of a simplex-valued vector in that it generates approximately indistinguishable outputs for inputs with such differences.
The Dirichlet mechanism takes as input a simplex-valued vector and a parameter $k$, and outputs an element of the simplex. 
Formally, for $\eta>0$ and $\pi\in\Delta_{n,\eta}$, the Dirichlet mechanism with input $\pi$ and parameter $k>0$, denoted $\mathrm{Dir}_k(\pi)$, outputs $x\in\Delta_m$ with probability
\begin{equation}
    \mathbb{P}\left(\mathrm{Dir}_k(\pi) = x\right) = \Gamma(k)\prod_{i\in[n]} \frac{x_i^{k\pi_i - 1}}{\Gamma(k(\pi_i))},
\end{equation}
where $\Gamma$ is the gamma function, defined as $\Gamma(z) = \int_0^\infty x^{z-1}\exp(-x)\mathrm{d}x$. The expected value of the Dirichlet mechanism's output is equal to its input for all values of $k$.
The parameter $k$ adjusts the concentration of the output around its expected value and the distribution becomes more concentrated around the input as $k$ increases.

The Dirichlet mechanism may achieve a higher accuracy than the conventional differential-privacy mechanisms because it does not require additional projections back onto the unit simplex.
For example, in comparison with a Gaussian mechanism that provides the same level of $(\epsilon,\delta)$-differential privacy, the empirical mean of the 1-norm distance between the input and output of the Dirichlet mechanism has been found to be almost $50\%$ less than that of the Gaussian mechanism \cite{9328150}.

The adjacency relationship upon which the Dirichlet mechanism was originally defined labels two simplex-valued vectors adjacent if they differ in at most two of their components with their 1-norm distance not exceeding a constant $b$.
We define the adjacency relationship differently and in accordance with the nature of the demonstrations that the teachers share with the students in kickstarting.
Hence, we need to show that the Dirichlet mechanism enforces $(\epsilon,\delta)$-differential privacy with respect to the adjacency relationship in Definition \ref{def:adjacency}. 
To this end, we will first generalize the adjacency relationship in \cite{gohari2020privacy} to one that applies to simplex-valued vectors whose 2-norm distance is below a constant $b$, across all of their components.
Then, we use the property of Lipschitz continuity of neural networks to prove differential privacy with respect to the adjacency relationship that we established in Definition \ref{def:adjacency}.

Neural networks with conventional activation functions such as $\mathrm{ReLU}$, $\mathrm{sigmoid}$, and $\mathrm{softmax}$ are known to be Lipschitz continuous \cite{combettes2019lipschitz,42503}.
Recall that a function $f:\mathbb{R}^n\mapsto\mathbb{R}^m$ is $L$-Lipschitz continuous if and only if there exists a constant $L\in\mathbb{R}_+$ such that, for all $x$ and $y$ in $\mathbb{R}^n$, $\|f(x) - f(y)\|_2 \le L \|x-y\|_2$.
The Lipschitz constant of a neural network can be computed efficiently due to an algorithm by \cite{fazlyab2019efficient}.
We use the Lipschitz coefficient of the teacher's neural network to establish an upper bound on the distance between the policies of two $b$-adjacent environment observations.
Then, we show that the Dirichlet mechanism can generate statistically similar outputs for such policies, and by that, we show that the Dirichlet mechanism satisfies the requirements of $(\epsilon,\delta)$-differential privacy established in Definition \ref{def:differential privacy}.
We now state the main theorem of this section.

\begin{theorem} \label{thm:differential privacy}
Fix an $L$-Lipschitz function $f:\mathbb{R}^d\mapsto\Delta_{m,\eta}$ and a Dirichlet mechanism $\mathrm{Dir}_k$ for some parameter $k>0$. Then, the composition mechanism $M = \mathrm{Dir}_k \circ f$ is $(\epsilon,\delta)$-differentially private with
\begin{equation} \label{eq:delta}
    \delta = 1 - \mathbb{P}\left(\mathrm{Dir}_k\left([\eta, \ \eta, \ \dots, \ 1-(m-1)\eta]^\top\right) \in \Delta_{m,\tau} \right)
\end{equation}
and
\begin{equation} \label{eq:epsilon}
    \epsilon = \mathcal{O}\left(k\log(k) \cdot b \cdot L \cdot \log\left(\frac{1}{\tau}\right)\right),
\end{equation}
where $b$ is the adjacency-relationship coefficient and $\tau>0$. 
\begin{proof}
We will show that the composition mechanism $M$ is $(\epsilon, \delta)$-differentially private by showing that it satisfies $(\epsilon,\delta)$-probabilistic differential privacy. Fix a constant $\tau\in\mathbb{R}_+$ and let $R_1 = \Delta_m \setminus \Delta_{m,\tau}$.
We must find $\delta$ such that, for all inputs $X$ in the domain of $M$ and all measurable subsets $S\subseteq R_1$, $\mathbb{P}\left(M(X)\in S\right)\le \delta$.
By Lemma 2 of \cite{9328150}, the function $I_{k,\tau}:\Delta_{m,\eta}\mapsto\mathbb{R}$ with
\begin{equation} 
    I_{k,\tau}(p) = \int_{\Delta_{m,\tau}} \Gamma(k)\prod_{i\in[m]} \frac{x_i^{kp_i - 1}}{\Gamma(k(p_i))} \mathrm{d}x
\end{equation}
is log-concave and its input domain is a convex set; therefore, its minimum occurs at the extreme points of its domain space $\Delta_{m,\eta}$. The extreme points of $\Delta_{m,\eta}$ are the basic feasible solutions of the system of $m$ inequalities
\begin{equation}\label{eq:system of inequalities}
    \forall i\in[m-1]: x_i \ge \eta, \quad \text{and } \sum_{i\in[m-1]} x_i \le 1-(1-m)\eta.
\end{equation}
A basic feasible solution of $m$ inequalities in $\mathbb{R}^n$ must activate $n$ of the inequalities and, in the inequality system in (\ref{eq:system of inequalities}), we have $n=m-1$. Therefore, the vertices of $\Delta_{m,\eta}$ have $m-1$ components equal to $\eta$ and one component equal to $1-(m-1)\eta$. Because $I_{k,\tau}$ is symmetric in the components of its input, for all $p\in\Delta_{m,\eta}$, $I_{k,\tau}(p) \ge I_{k,\tau}(p^*)$, where $p^*$ is any vertex of $\Delta_{m,\eta}$. 
Note that
\begin{equation} \label{eq:thm1 0}
    I_{k,\tau}(p) = \mathbb{P}\left(\mathrm{Dir}_k\left(p\right) \in \Delta_{m,\tau} \right), \quad \forall p\in\Delta_{m,\eta}.
\end{equation}
Any distribution according to which the input $X\in\mathbb{R}^d$ is drawn satisfies
\begin{equation}
    \mathbb{P}\left(M(X)\in S\right) \le \max_p \{1 - \mathbb{P}\left( \mathrm{Dir}_k(p) \ge \tau \right)\}
    = 1 - \min_p I_{k,\tau}(p) = 1 - I_{k,\tau}(p^*), \quad \forall S\subseteq R_1.
\end{equation}
Using (\ref{eq:thm1 0}), we can conclude (\ref{eq:delta}).
To show that \eqref{eq:epsilon} holds, in the supplementary materials, we show that, for all $\pi$ and $\pi'$ in $\Delta_{m,\eta}$ such that $\|\pi-\pi'\|_2 \le b$ and all $z\in R_2 = \Delta_{m,\tau}$,
\begin{multline*}
    \log\left({\mathbb{P}(\mathrm{Dir}_k(\pi) = z)}\right) \le \log\left(\mathbb{P}(\mathrm{Dir}_k(\pi') = z)\right)
    +\sqrt{m}bk|\log(\tau)| \\
    +(m-1) \log(\Gamma(k\eta)) + \log(\Gamma(k(1-(m-1)\eta))) 
    - m \log(\Gamma(k/m)).
\end{multline*}
Then, using the fact that $f$ is $L$-Lipschitz continuous, we find $\epsilon$ in \eqref{eq: prob DP epsilon} as
\begin{multline}
    \epsilon = \sqrt{m}Lbk\log(1/\tau) + (m-1) \log(\Gamma(k\eta)) + \log(\Gamma(k(1-(m-1)\eta))) 
    - m \log(\Gamma(k/m)),
\end{multline}
and we conclude the theorem.
\end{proof}
\end{theorem}

\textbf{Remark.} Computing $\delta$ in (\ref{eq:delta}) requires computing an $m$-fold integral of the Dirichlet probability distribution function over the $\eta$-restricted simplex, and may be computationally expensive. In the implementations, we approximately compute $\delta$ for fixed values of $\tau$. In particular, let $Z$ be a Bernoulli random variable that equals 1 if $\mathrm{Dir}_k\left([\eta, \ \eta, \ \dots, \ 1-(m-1)\eta]^\top\right) \not \in \Delta_{m,\tau}$, and equals 0 otherwise. Then, by the Chebyshev inequality, the empirical mean of $N$ observations of $Z$ satisfies
\begin{equation}
    \mathbb{P}\left(\left|\frac{1}{N}\sum_{i\in[N]} Z_i - \mathbb{E}[Z]\right| \ge t\right) \le \frac{\sigma_Z^2}{t^2N^2},
\end{equation}
where $\sigma_Z$ is the variance of $Z$ and the variance of Bernoulli variables is upper bounded by $1/4$. We note that $\mathbb{E}[Z] = 1 - \mathbb{P}(\mathrm{Dir}_k\left([\eta, \ \eta, \ \dots, \ 1-(m-1)\eta]^\top\right) \in \Delta_{m,\tau})$.
Hence, we can compute a $t$-approximation of $\delta$ by evaluating the output of the Dirichlet mechanism $\mathcal{O}(t^{-2})$ times.

\section{Privacy-Aware Students} \label{sec:student}
In this section, we complete the privacy-preserving kickstarting algorithm by designing the student's learning algorithm in accordance to the privacy mechanism that we developed in the previous section.
By following a teacher-student framework, we implicitly assume that the teacher's policy performs better than the student's policy; therefore, it is desirable for the student's policy to approximate the teacher's policy.
However, due to the perturbations of the privacy mechanism, the shared demonstration may mislead the student.
As a result, the student must be aware of the perturbations and rely on its own environment interactions when the perturbations are too strong.
The following concentration bound from \cite{gohari2020privacy} helps us determine the magnitude of the perturbation of the teacher's privacy mechanism.
\begin{lemma}\label{lemma:concentration}
Let $\mathrm{Dir}_k$ be a Dirichlet mechanism with parameter $k>0$, and $\pi\in\Delta_{n,\eta}$, for some $\eta>0$. Then, for all $\beta>0$,
\begin{equation} \label{eq:concentration}
    \mathbb{P}\left(\left\|\mathrm{Dir}_k(\pi) - \pi\right\|_2 \ge \sqrt{\frac{\log\left(1/\beta\right)}{2(k+1)}}\right) \le \beta.
\end{equation}
\end{lemma}
Lemma \ref{lemma:concentration} determines the probability of large deviations of the output of the Dirichlet mechanism from its input. 
The student only observes the output policy that the Dirichlet mechanism generates and, for a fixed $\beta$, a 2-norm ball around the observed policy is likely to contain the teacher's true policy.
Therefore, if the student attempts to fully approximate the observed policy, it is likely that it will be attempting to learn from noisy data which can be misleading.
As a result, we will only penalize the student's approximation as long as it lies outside the 2-norm ball in \ref{lemma:concentration}.

Similar to the existing kickstarting methods \cite{schmitt2018kickstarting, rusu2015policy}, we allow the student to simultaneously learn from the teacher demonstrations and from its interactions with the environment.
However, instead of using an information-theory measure, we use the 2-norm distance to augment the student's objective because it is simpler to implement and it is compatible with the concentration bound in Lemma \ref{lemma:concentration}.

We denote the student's parametric policy by $\pi_\theta(s)\in\Delta_{m}$, where $\theta$ represents the parameters of the student's underlying neural network and $s$ is a state observation in $\mathbb{R}^d$, and denote the teacher's privacy-preserving demonstration by $\pi^t(s)$.
Let $J(\theta)$ denote the objective function of the deep reinforcement learning algorithm upon which we base the student's algorithm.
The objective function $J$ usually represents the expected total rewards of the agent or some surrogate function thereof; however, its implementation varies across existing algorithms.

Let $k$ denote the parameter of the teacher's Dirichlet mechanism and fix a confidence level $\beta$.
We define the variable $\alpha_\lambda$ to capture the radius of the 2-norm ball in \eqref{eq:concentration} as follows:
\begin{equation} 
    \alpha_\lambda = \lambda \cdot \sqrt{\frac{\log(1/\beta)}{2(k+1)}},
\end{equation}
where $\lambda$ is a hyperparameter by which we can shrink the ball if the bound is conservative.
We then define the function
\begin{equation} \label{eq:phi}
    \phi_\lambda\left(s,\pi^{(t)}\right) := \left\|\pi_\theta(s) - \pi^\text{t}(s)\right\|_2 \cdot \mathcal{I}_{\left\{\left\|\pi_\theta(s)- \pi^\text{t}(s)\right\|_2 > \alpha_\lambda\right\}},
\end{equation}
where $\mathcal{I}$ is the indicator function that activates the teacher demonstrations and
\begin{equation*}
\mathcal{I}_{\left\{\left\|\pi_\theta(s)- \pi^\text{t}(s)\right\|_2 \ge \alpha_\lambda\right\}} := 
\begin{cases}
1 & \ \  \text{if}  \left\|\pi_\theta(s)- \pi^\text{t}(s)\right\|_2 \ge \alpha_\lambda,\\
0 & \ \  \text{otherwise}.
\end{cases}
\end{equation*}
Finally, we propose the augmented objective function
\begin{equation} \label{eq:objective}
    \tilde J(\theta) = J(\theta) - \mathbb{E}_{s\sim d^{\pi_\theta}, \pi^\text{t} \sim \mathrm{Dir}_k}\left[ \phi_\lambda(s,\pi^\text{t})\right],
\end{equation}
where $d^{\pi_\theta}$ is the stationary state distribution under the student's policy. In case the two terms on the right-hand side of \eqref{eq:objective} have unbalanced orders of magnitude, we can introduce an additional hyperparameter to balance the two terms. However, we did not include the additional hyperparameter to keep the notations simple.

The objective function in \eqref{eq:objective} makes the student's learning algorithm privacy-aware because it takes the perturbations of the privacy mechanism into account. Moreover, by maximizing $\tilde J(\theta)$, the student will not blindly copy the teacher's demonstrations and remains a deep reinforcement learning agent.


Popular reinforcement learning algorithms such as trust-region policy optimization \cite{schulman2017proximal} and proximal policy optimization \cite{schulman2015trust} optimize some surrogate function of the objective $J(\theta) = \mathbb{E}_{s\sim d^{\pi_\theta}} \left[V^{\pi_{\theta}}(s)\right]$, where $V^\pi$ is the value function of the agent's policy.
To approximate the expectation above, the algorithms use roll-out trajectories to compute the empirical mean of the discounted total reward of the policy.
Similar to the these algorithms, we compute the expectation $\mathbb{E}_{s\sim d^{\pi_\theta}, \pi^\text{t} \sim \mathrm{Dir}_k}\left[ \phi_\lambda(s,\pi^\text{t})\right]$ from roll-out trajectories.

\textbf{Remark.} We are mindful of the fact that subsequent queries from a differential privacy mechanism may weaken its strength of privacy protections.
In particular, the composition of $k$ mechanisms with each mechanism enforcing $(\epsilon,\delta)$-differential privacy mechanism will enforce an overall differential privacy of $(k\epsilon,k\delta)$, given that the same input is queried \cite{dwork2014algorithmic}.
Therefore, it is conventional to allocate a privacy budget for repeated queries in differential privacy \cite{abadi2016deep,dankar2013practicing}.
In kickstarting, the teachers do not answer the same query over and over again; however, the sequence of the student's environment observations are correlated via the dynamics of the environment and we must allow the teachers to allocate and manage a privacy budget for themselves.

To allow the teachers to manage their privacy budget, we decrease the concentration parameter of the Dirichlet mechanism, $k$, at the start of each roll-out trajectory. A decrease in $k$ both strengthens the Dirichlet mechanism's differential privacy and decreases the accuracy of the output policy.
Depending on the rate at which we decrease $k$, at a certain point, the objective of the student will completely disregard the teacher demonstrations by \eqref{eq:phi} and the student will rely on its own interactions with the environment.
At that point, the teachers can generate completely random policies under $(0,0)$-differential privacy without harming the student's performance.
Therefore, the teachers can decrease $k$ according to their privacy budget and set $k=0$ once they expend their budget.
However, teachers with higher privacy budgets may be more beneficial to the students, which is indeed what we observe in the next section.



\section{Numerical Results} \label{sec:exp}
In this section, we showcase the performance of the privacy-preserving kickstarting deep reinforcement learning algorithm that we developed in the previous two sections.
We implement the algorithm using the Spinning Up library \cite{SpinningUp2018} and the experiments take place in the \textit{AirRaid-ram-v0} environment from OpenAI Gym \cite{1606.01540}. All of the reinforcement learning agents use proximal policy optimization (PPO) as their base algorithm. In the sequel, we perform four experiments. The first experiment establishes our choice of teacher and student agents. In the second experiment, we deploy kickstarting for the agents without any privacy protections. In the third experiment, we test the privacy-preserving kickstarting algorithm using privacy-aware students, which is the main contribution of this paper. In the last experiment, we show that privacy-awareness is necessary for the students to retain the benefits of kickstarting while protecting the privacy of the teachers.

\begin{figure}
    \centering
    \includegraphics[width=0.94\textwidth, height = 4.4in]{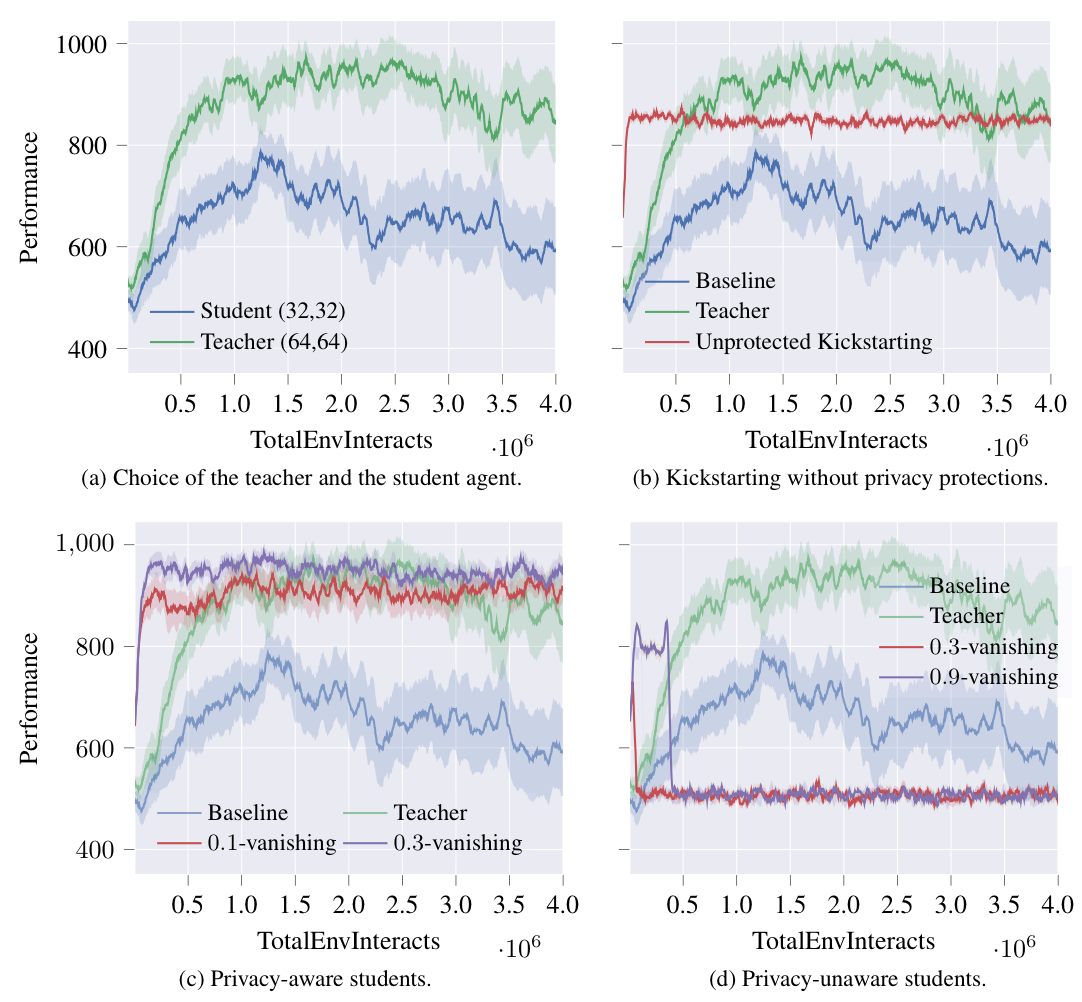}
    \caption{Showcasing the effectiveness of privacy-preserving kickstarting deep reinforcement learning in training agents in OpenAI Gym's \textit{AirRaid-ram-v0} environment.}
    \label{fig:1}
\end{figure}

In the first experiment, we train two agents with different neural network sizes. The first agent has two hidden layers, each of which has 64 neurons and the second agent has two hidden layers with 32 neurons within each layer. Figure \ref{fig:1}(a) shows the learning curves of the two agents across four million environment interactions and indicates that the agent with a smaller hidden layer performs worse than the other agent. Therefore, we use the converging policy of the better performing agent as the teacher to kickstart the training of the other agent.


In the second experiment, we use kickstarting to train the student but we do not deploy the Dirichlet mechanism yet. Because of the absence of perturbations, we set $\lambda$ in \eqref{eq:objective} to zero on the student's side.
As a result, the algorithm reduces to conventional kickstarting without any privacy protections.
The results are shown in Figure \ref{fig:1}(b) and they are indicative of the known benefits of kickstarting, namely a significant improvement in the students' sample efficiency \cite{schmitt2018kickstarting} and \textit{model compression}. Model compression methods, such as policy distillation \cite{rusu2015policy}, employ teacher agents with sophisticated models to train agents with smaller networks that approximate the performance level of the teachers. As the final observation, the run-to-run variance of the kickstarting algorithm is significantly lower than the teacher and the student who does not benefit from kickstarting. Therefore, the students can improve upon their robustness under kickstarting.

In the third experiment, we deploy the Dirichlet mechanism. 
To manage the privacy budget of the teacher, we follow the remark at the end of Section 4 and progressively increase the level of the teacher's differential privacy until they no longer carry any useful information. 
Specifically, we multiply the Dirichlet mechanism's parameter $k$ by a vanishing coefficient at the beginning of each roll-out trajectory. 
Figure \ref{fig:1}(c) compares the performance of two privacy-aware students under privacy-preserving kickstarting with vanishing coefficients $0.1$ and $0.3$. 
Both agents converge roughly 4 times faster than the teacher; 
however, kickstarting with a lower vanishing coefficient consumes a lower privacy budget and the resulting student has a lower performance.
The student's performance corresponding to a lower privacy budget is also less robust to environmental changes than those with higher budgets as the confidence clouds suggest in Figure \ref{fig:1}(c). 

Figure \ref{fig:1}(c) corresponds to $\lambda = 0.5$. In the supplementary materials, we show the performance of the algorithm for $\lambda \in \{0.1, 0.5, 1, 2\}$ and choose $\lambda=0.5$ because it is associated with the student's best performance.
We also provide additional plots in the supplementary materials that correspond to the vanishing coefficients in $\{0.1,0.2,0.3,0.4,0.5,0.6,0.7,0.8,0.9\}$.

The fourth experiment demonstrates that it is necessary for the students' learning algorithms to be privacy-aware when the the teachers deploy their privacy mechanisms. We set $\lambda$ to zero in \eqref{eq:objective} and compare the performance of two students under privacy-preserving kickstarting with vanishing coefficients $0.3$ and $0.9$ in Figure \ref{fig:1}(d). In comparison with the results in Figure \ref{fig:1}(c), the privacy-unaware students perform significantly worse. In fact, once the teachers expend their privacy budget and share completely random policies, the students perform worse than the baseline PPO agents that do not use kickstarting at all. 

In all of the experiments above, the confidence clouds correspond to the 85th percentile of 30 random seeds of the environment, and the lines represent to the point-wise median of the runs.
In the experiments where the Dirichlet mechanism is deployed, we initialize $k$ with $5$ which corresponds to $(4.46,0.05)$-differential privacy by Theorem \ref{thm:differential privacy}.
Unless specified, for example, for the network size, we have used the default initialization of the Spinning Up library to set the hyperparameters of the PPO algorithm.
The simulations were run on a computer with a $3.5$ GHz quad-core Intel processor. The longest computation time per epoch was 9 seconds where every epoch consists of 1000 environment interactions.
All codes and datasets are included in the supplementary materials.


\section{Conclusions and Discussions} \label{sec:limitations}
In this paper, we developed a privacy-preserving kickstarting algorithm as a result of two main contributions: first, we designed a privacy mechanism that protects the differential privacy of the teachers, and secondly, we designed a privacy-aware learning algorithm for the students that can benefit from the privacy-preserving demonstrations that the teachers share with them.

By Theorem \ref{thm:differential privacy}, we show that the policy that the Dirichlet mechanism outputs is probable to be the output of a class of environment observations that are adjacent by Definition \ref{def:adjacency}.
In the example of membership attacks, the attackers train \textit{shadow} models that copy the behavior of their victim agents and use these shadow models to answer the question "Has a given environment observation been used to train the agent or not?"
Theorem 1, guarantees that, if the membership attacker answers 'yes' to a specific environment observation, it is probable that it answers 'yes' to every other adjacent environment observation. Therefore, the attacker cannot recover the exact training data of the teacher by observing its inputs and outputs.

Although we impose an error on recovering the exact training data, the attacker can still recover an approximation of the training data and, depending on the privacy budget that the teacher consumes, the attacker can improve its approximation by repeating its query.
Therefore, the potential users of the privacy-preserving kickstarting algorithm that we propose must be mindful of the nature of the privacy protections that we offer to avoid any privacy breaches.
As a result, a possible negative consequence of this paper is that the privacy protections that we provide may not align with users’ intuition for privacy.

\section{Related Works}
In this section, we review some of the existing works that protect the user privacy of machine learning algorithms.
The work in \cite{10.1145/2976749.2978318} develops a privacy-preserving deep learning algorithm that enforces differential privacy by injecting a Gaussian noise to the gradients that are used to update the parameters of the neural network. The works in \cite{chaudhuri2011differentially, zhang2012functional, jayaraman2018distributed} enforce differential privacy for a classifier by injecting noise to its objective function. 
These methods require access to the internal models of the teachers and are invasive. In this paper, however, we do not require access to the internal model of the teachers because we privatize the policy, which is the output of the teachers' neural networks.

The work in \cite{papernot2016semi} considers the problem of knowledge transfer between a collection of teachers and a student, which is closely related to this paper. To enforce differential privacy, the authors of \cite{papernot2016semi} perform a randomized voting between the teachers to choose the teacher that shares the demonstration. The requirement of voting makes the methodology not applicable to problems with one teacher, whereas in this paper, we can consider individual teachers.

\section{Appendix}

For all $\pi$ and $\pi'$ in $\Delta_{m,\eta}$ such that $\|\pi-\pi'\|_2 \le b$ and all $z\in R_2 = \Delta_{m,\tau}$, we write
\begin{equation}
    \log\left(\frac{\mathbb{P}(\mathrm{Dir}_k(\pi) = z)}{\Gamma(k)}\right) = \sum_{i\in[n]}(k\pi_i - 1) \log(z_i)    - \sum_{i\in[n]}{\log(\Gamma(k(\pi_i)))}.
\end{equation}
We then write
\begin{multline*}
    \log\left({\mathbb{P}(\mathrm{Dir}_k(\pi) = z)}\right) = \log\left(\mathbb{P}(\mathrm{Dir}_k(\pi') = z)\right)
    +\sum_{i\in[m]}(k(\pi_i-\pi'_i))\log(z_i) \\
    -\sum_{i\in[m]} \log(\left(\Gamma\left(k\pi_i\right)\right)+\sum_{i\in[m]}\log\left(\Gamma\left(k\pi'_i\right)\right).
\end{multline*}

Using the fact that $z\in R_2$, we have that, for all $\|\pi-\pi'\|_2 \le b$ and small values of $\tau$,
\begin{equation}\label{eq:theorem 1 , i}
    \sum_{i\in[m]}(k(\pi_i-\pi'_i))\log(z_i) \le \sqrt{m}bk|\log(\tau)|.
\end{equation}

We now show that 
\begin{multline}
    -\sum_{i\in[m]} \log(\left(\Gamma\left(k\pi_i\right)\right)+\sum_{i\in[m]}\log\left(\Gamma\left(k\pi'_i\right)\right) \le \\
     (m-1) \log(\Gamma(k\eta)) + \log(\Gamma(k(1-(m-1)\eta))) 
    - m \log(\Gamma(k/m)).
\end{multline}

The composition of the logarithm and the gamma function is convex \cite{weisstein2005log}. 
Evaluating the Karush-Kuhn-Tucker optimality conditions corresponding to the optimization problem
\begin{align}
    &\max_{\pi} & &-\sum_{i\in [m-1]}\log(\Gamma(k\pi_i)) -  \log\left(\Gamma\left(k\pi_m\right)\right) & & \\
    &\text{s.t.} & &\forall i\in[m-1]: \pi_i \ge \eta \nonumber\\
    & & & \pi_m = 1-\sum_{i\in [m-1]}\pi_i, \nonumber
\end{align}
the gradient of the objective function, which is concave, at $\pi = [k/m, k/m, \dots, k/m]$ is zero and $\pi$ is an interior point of $\Delta_{m,\eta}$. 
Therefore,
\begin{equation}\label{first}
    -\sum_{i\in[m]} \log(\left(\Gamma\left(k\pi_i\right)\right) \le - m \log(\Gamma(k/m)).
\end{equation}

We now consider the optimization problem
\begin{align}
    &\max_{\pi} & &\sum_{i\in [m-1]}\log(\Gamma(k\pi_i)) +  \log\left(\Gamma\left(k\pi_m\right)\right) & & \\
    &\text{s.t.} & &\forall i\in[m-1]: \pi_i \ge \eta \nonumber\\
    & & & \pi_m = 1-\sum_{i\in [m-1]}\pi_i. \nonumber
\end{align}
The objective above is convex and the space of the feasible solutions is convex; therefore, the maximum occurs at an extreme point. As a result,
\begin{equation}\label{second}
    \sum_{i\in[m]}\log\left(\Gamma\left(k\pi'_i\right)\right) \le (m-1) \log(\Gamma(k\eta)) + \log(\Gamma(k(1-(m-1)\eta))).
\end{equation}
Combining \eqref{first}, \eqref{second},  we have that that 
\begin{multline}
    -\sum_{i\in[m]} \log(\left(\Gamma\left(k\pi_i\right)\right)+\sum_{i\in[m]}\log\left(\Gamma\left(k\pi'_i\right)\right) \le \\
     (m-1) \log(\Gamma(k\eta)) + \log(\Gamma(k(1-(m-1)\eta))) 
    - m \log(\Gamma(k/m)).
\end{multline}

\end{document}